\documentclass{article}
\usepackage{ijcai19}

\usepackage{times}
\usepackage{soul}
\usepackage{url}
\usepackage[hidelinks]{hyperref}
\usepackage[utf8]{inputenc}
\usepackage[small]{caption}
\usepackage{graphicx}
\usepackage{subfigure}
\usepackage{amssymb}
\usepackage{amsmath}
\usepackage{booktabs}
\usepackage{algorithm}
\usepackage{algorithmic}
\usepackage{CJK}
\usepackage{color}
\title{Image Captioning based on Deep Learning Methods: A Survey}

\author{
Yiyu Wang\footnote{Y. Wang and J. Xu are corresponding authors}
\and 
Jungang Xu$^*$
\and
Yingfei Sun
\And
Ben He
\affiliations
University of Chinese Academy of Sciences, Beijing\\
\emails
wangyiyu18@mails.ucas.ac.cn,
\{xujg,yfsun,benhe\}@ucas.ac.cn
}
\begin{document}
\begin{CJK}{UTF8}{gbsn}

\maketitle

\begin{abstract}
  Image captioning is a challenging task and attracting more and more attention in the field of Artificial Intelligence, and which can be applied to efficient image retrieval, intelligent blind guidance and human-computer interaction, etc. In this paper, we present a survey on advances in image captioning based on Deep Learning methods, including Encoder-Decoder structure, improved methods in Encoder, improved methods in Decoder, and other improvements. Furthermore, we discussed future research directions.
\end{abstract}

\section{Introduction}
There are a large number of unlabeled images in the network; it is impossible to label them manually. How to automatically generate natural language descriptions for images by computer is a challenging task in the field of artificial intelligence. Image captioning can be applied to efficient image retrieval, intelligent blind guidance and human-computer interaction, so it is also a task with practical value.

The goal of image captioning is to generate a trusted description for a given image. So, it is necessary to ensure the correctness of the objects, attribute information, semantic information, and position relationship information in the description. Therefore, we can decompose image captioning into two sub-tasks: (1) understanding the image, acquiring the relevant information correctly; (2) generating description based on the understanding of the image. Image captioning is a challenging task because it connects the two fields of Computer Vision(CV) and Natural Language Processing(NLP).

In other words, image understanding is equivalent to feature extraction. In traditional methods, the bottom visual features (such as geometry, texture, colour, etc.) are extracted by using artificially designed feature operators, and then combined to form high-level global features. However, there are some drawbacks in these traditional methods. On one hand, the design of feature operator relies too much on luck and experience. On the other hand, the problem of "semantic gap" leads to the inability of low-level visual features to accurately express semantic features. Therefore, traditional methods lack robustness and generalisation performance.

For a given image, the retrieval-based method selects sentence(s) from a specified image-description pool as the description(s) of the image; the template-based method detects a series of specified visual features from the image, and then fills them into the blank position of the given template. Images are very complex data. The description extracted by the retrieval-based method may not fully conform to the image. The image description generated by template-based method seems too rigid and lacks diversity.

In recent years, Convolutional Neural Networks (CNN) have obtained outstanding effects in CV tasks, such as image classification, object detection. Recurrent Neural Networks (RNN) also played a significant role in NLP. In addition, inspired by Encoder-Decoder structure in machine translation \cite{DBLP:conf/nips/SutskeverVL14}, \cite{DBLP:conf/cvpr/VinyalsTBE15} uses GoogLeNet as Encoder to automatically extract image features, and then uses Long and Short-Term Memory network (LSTM) \cite{DBLP:journals/neco/HochreiterS97} as Decoder to generate description, which is a pioneering work of image captioning using deep learning. Since then, Deep Learning methods based on Encoder-Decoder structure have become the basic framework of image captioning.

In the past few years, a large number of research works based on Deep Learning methods were published. Many useful improvements are proposed based on Encoder-Decoder structure, such as semantic attention \cite{you_image_2016}, visual sentinel \cite{lu_knowing_2017}, and review network \cite{yang_review_2016}. We divide them into (1) Improvements in Encoder (2) Improvements in Decoder and (3) Other Improvements. 

The main contributions of this paper include:(1) introduced and analyzed traditional methods such as Retrieval-Based and Template-Based methods; (2) provided an overview of Encoder-Decoder structure; (3)summarized improvements in Encoder and Decoder for image captioning; (4)discussed and proposed future research directions. 

The rest of this paper is organized as follows. Section 2 introduces the traditional image captioning methods. Section 3 focuses on the improvements in Encoder-Decoder. Section 4 and 5 introduce the existing standard Datasets and evaluation metrics. Section 6 discusses the future research directions. Section 7 gives the conclusions.

\section{Traditional Methods}
%Previous works on image captioning largely adopt retrieval-based and template-based methods. 
This paper mainly focuses on deep learning methods. Hence, in this part, we only briefly review retrieval-based and template-based methods as traditional methods.
\begin{figure}[]
    \centering
    \subfigure{\includegraphics[width=3.2in]{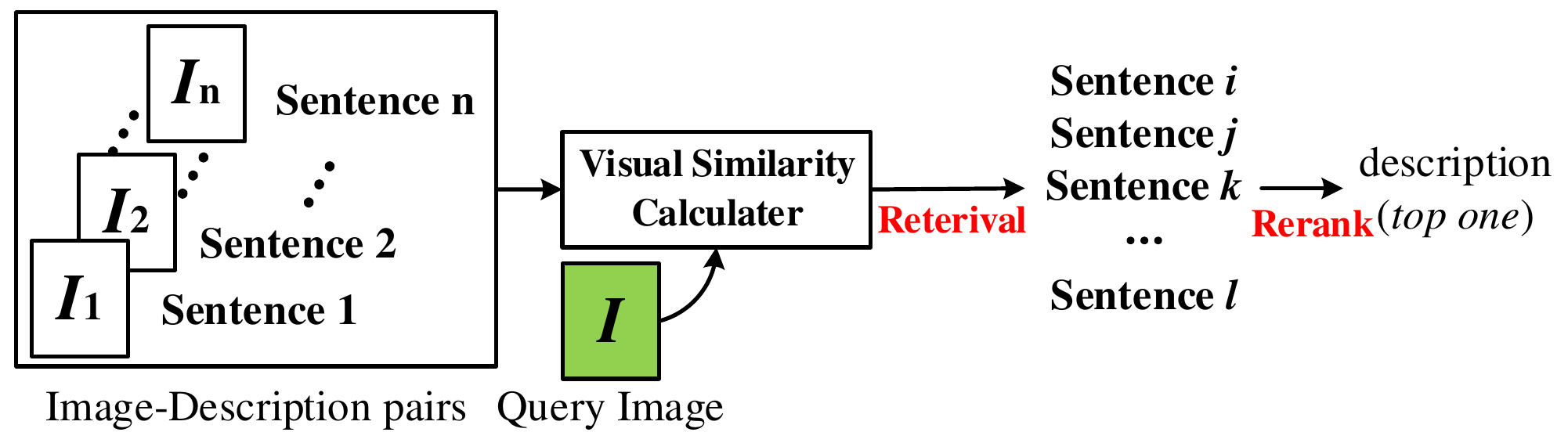}}
    \subfigure{\includegraphics[width=3.2in]{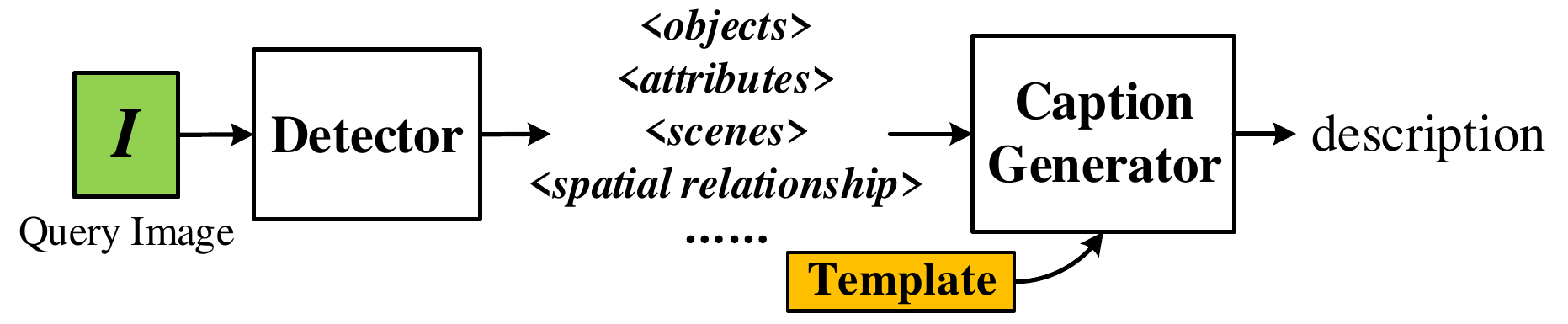}}
    \caption{The fundamental process of traditional methods: Reterival-Based Method (top). Template-Based Method (bottom).}
\end{figure}

\subsection{Retrieval-Based Methods}
For a given image, the retrieval-based image captioning methods aim to retrieve the matching sentence(s) from a set of given image-description pairs as the language description of the image, see Figure 1 (top). Therefore, the quality of this method depends on not only the diversity of image-description pairs but also the image-based retrieval algorithm.

\cite{DBLP:conf/nips/OrdonezKB11} firstly retrieves a series of related images from the image-description pairs by Gist and Tiny image descriptors, then detects and classifies the query images by specific objects and scenes, and reorders the retrieved images in turn, choosing the description of the first image to be ranked as the description of the query image. This method can be regarded as retrieval in visual space.

%\cite{DBLP:conf/acl/MasonC14} first applies visual similarity to retrieve a series of related images from the image-description pairs, then establishes the word probability density from the description of the retrieval image based on the query image, and rerank the retrieval image description based on the established word probability density. The top rank is the description of the query image. Above two methods can be regarded as retrieval in the visual space.

\cite{hodosh_framing_2013} regards image captioning as a ranking task, and KCCA (A kernelized version of canonical correlation analysis) is employed to project images and descriptions into a common multimodal space. Then the query image is projected into the multimode space, and the Cosine similarity between the query image and the descriptions in datasets are calculated. The top rank is accepted as the description of the query image. However, KCCA is only suitable for small datasets, which can affect the performance of this method. This method can be regarded as retrieval in multimodal space.

But, the shortcomings of the retrieval-based method are also explicit. The quality of the description generated by this method depends extensively on the given image-description pool. The image-description pairs are established artificially, so it is sufficient to ensure the fluency of the description sentence and the accuracy of the grammar; however, to ensure the accuracy of the description content and semantics, the pre-given image-description pairs need to be large sufficient to cover enough rich scenes. The limitation of this method may not suit the object and scene of new images correctly, so it also limits the generalisation performance of this method.

\subsection{Template-Based Methods}
For a given image, the template-based image captioning method usually requires to extract some objects, attributes or semantic information from the image, and then uses a specified grammar rule to combine the information or fills the obtained data into the pre-defined blanks of the sentence template to form the image description, see Figure 1 (bottom).

\cite{DBLP:conf/conll/LiKBBC11} firstly uses an image recogniser to obtain visual information from the image, including objects, attributes of objects and spatial relationships between different objects. And then they encode the information as a triple form of $<<$adj1, obj1 $>$, prep, $<$adj2, obj2 $>>$. Furthermore, an approach based on web-scale n-gram is used to get the frequency counts of all possible n-gram sequences ($1 \leq n \leq 5 $). Finally, the phrases are selected and fused, and the best combination is accepted as the description of the query image by the dynamic programming algorithm.

\cite{DBLP:conf/cvpr/KulkarniPDLCBB11} uses an object detector to detect objects in the image, and then sends candidate object regions into attribute classifier and prepositional relation function to obtain attribute information of candidate objects and prepositional relation information between objects. Furthermore, a Conditional Random Field (CRF) is constructed to deduce the relevant information previously obtained for final use. 

%\cite{DBLP:conf/emnlp/YangTDA11} uses a Nouns - Verbs - Scenes - Prepositions as a description template. The model first uses the detection model to obtain objects and scene information of the image. At the same time, in order to predict actions information from the image, the method uses a language model trained on the English Gigaword corpus to estimate the action, and calculates the probability of nouns, scenes and prepositions that constitute the sentence at the same time, and uses these estimates to generate the description of the image using HMM.

Compared with retrieval-based methods, template-based methods can also generate grammatically correct description statements, and because this method needs to detect objects from the image, the generated description is more relevant to the image to some extent. But, the deficiencies of template-based methods are also apparent. On one hand, sentence templates or grammar rules need to be pre-designed artificially, so this method can not generate variable-length sentences, which limits the diversity of descriptions between different images, and descriptions may seem rigid and unnatural; On the other hand, the performance of the object detector limits the accuracy of image description, so the generated description may omit the details of the query image.

\section{Deep Learning Methods}
In recent years, deep learning methods have made significant progress in CV and NLP. Inspired by machine translation \cite{DBLP:conf/nips/SutskeverVL14}, the Encoder-Decoder structure is also applied to image captioning. Usually, CNN is used to construct an Encoder to extract and encode information from images. RNN is used to construct a Decoder to generate descriptions. On this basis, many researchers have also proposed various efficient improvement methods, but they have different focuses. Therefore, we divide them into multiple sub-categories according to the improved focus, then introduce and discuss each sub-category separately.

\subsection{Basic Encoder-Decoder structure}
Show and Tell \cite{DBLP:conf/cvpr/VinyalsTBE15} is the first work to apply the Encoder-Decoder structure proposed in machine translation to image captioning. It also serves as the basis for subsequent improvements and a baseline model for performance comparison between models. The model structure is shown in Figure 2 (top).
\begin{figure}[]
    \centering
    \subfigure{\includegraphics[width=3.2in]{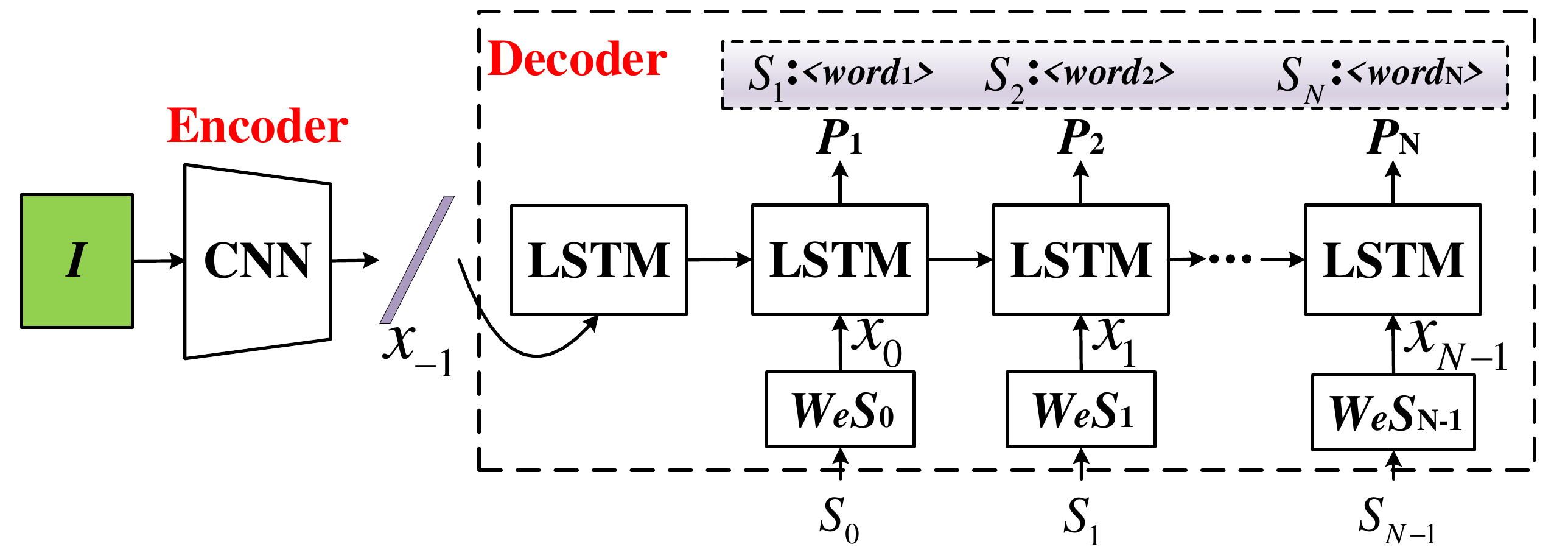}}
    \subfigure{\includegraphics[width=3.2in]{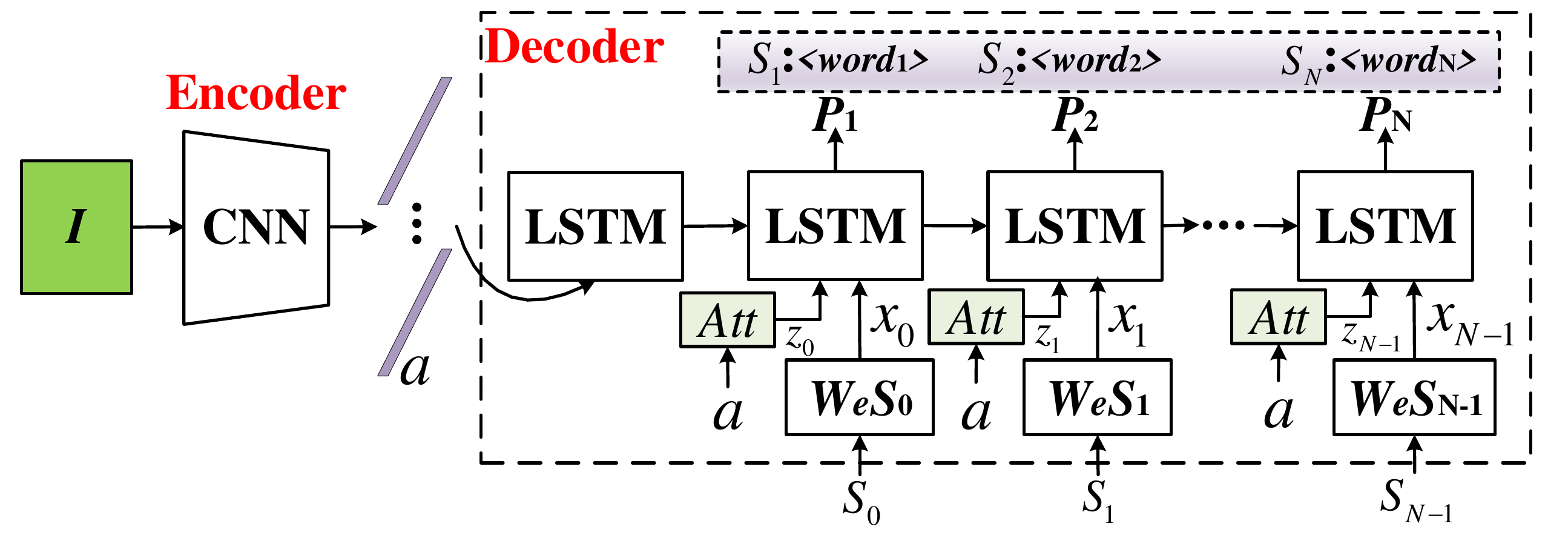}}
    \caption{Models Based on Encoder-Decoder structure: Show and tell (top). Show, attend and tell (bottom)}
\end{figure}

This model first uses the CNN as the Encoder part, encodes the image into a fixed-length vector representation as the image feature map, and then sends the image feature map to the Decoder part of the RNN to decode and generate an image description. It can be expressed as Eq.(1)-Eq.(3). The Encoder part is a CNN, which corresponds to GoogLeNet (Inception V3); the Decoder part is LSTM. 
\begin{align}
x_{-1}&={\rm Encoder}(I)\\
x_t&=W_eS_t, &t\in\{0,...,N-1\}\\
p_{t+1}&={\rm Decoder}(x_t), &t\in\{0,...,N-1\}
\end{align}

Suppose the vocabulary size is $D$, where $I$ represents the input image, $x_{-1}$ is the feature map, which is only used to initialize the LSTM; $S_t$ is the one-hot vector in size $D$, representing the $t$-th word of the image description, and $S_0$ is the $<$START$>$ tag, $S_N$ is the $<$END$>$ tag; $W_e$ is the word embedding matrix; $p_{t+1}\in \mathbb{R}^{D}$ represents the probability vector generated by the $t+1$ time step, wherein the most probable one corresponds to the word as the time step word output.

%The network structure is shown in Figure x and is defined as follows.
%\begin{figure}
%    \centering
%    \subfigure{\includegraphics[width=3in]{lstm.pdf}}
%    \caption{LSTM structure.}
%\end{figure}
%\begin{equation}
%\begin{split}
%i_t&=\sigma(W_{ix}x_t+W_{im}m_{t-1})\\
%f_t&=\sigma(W_{fx}x_t+W_{fm}m_{t-1})\\
%o_t&=\sigma(W_{ox}x_t+W_{om}m_{t-1})\\
%g_t&=\tanh(W_{gx}x_t+W_{gm}m_{t-1})\\
%c_t&=f_t\odot c_{t-1}+i_t\odot g_t\\
%m_t&=o_t\odot c_t\\
%p_{t+1}&={\rm Softmax}(m_t)
%\end{split}
%\end{equation}

Show Attend and Tell \cite{xu_show_2015} is an extension of \cite{DBLP:conf/cvpr/VinyalsTBE15}, which introduces a visual attention mechanism based on the Encoder-Decoder structure, which can dynamically focus on the salient regions of the image during the process of generating descriptions in Decoder. The model structure is shown in Figure 2 (bottom).

This model also uses a CNN as Encoder to extract $L$ vectors of $K$ dimensions from the image, each vector corresponds to a portion of the image. But unlike \cite{DBLP:conf/cvpr/VinyalsTBE15}, the model uses the underlying convolutional layer output instead of the final fully connected layer output as the image feature vector.
\begin{equation}
 a=\{a_1,...,a_L\}, a_i\in \mathbb{R}^K
\end{equation}

In the Decoder part, \cite{xu_show_2015} also uses LSTM for description generation. But this model needs to use the image-based feature vector $a$ for each time step $t$ to generate the context vector $z_t=\sum_{i=1}^{L } \alpha_{ti}a_i$. This is the embodiment of the attention mechanism, $\alpha_t\in \mathbb{R}^L$ is the attention weight vector of the $t$ time step, which satisfies $\sum_{i=1}^{L }\alpha_{ti}=1$. $a$ can be predicted by the simple neural network $f_{\rm att}$ and the Softmax activation function.
\begin{equation}
\alpha_{ti}\propto \exp\{f_{\rm att}(a_i,m_{t-1})\}
\end{equation}

Therefore,  the attention Encoder-Decoder structure can be expressed as Eq.(6)-Eq.(9).
\begin{align}
a&={\rm Encoder}(I)\\
z_t&=\sum_{i=1}^{L} \alpha_{ti}a_i, &&\alpha_{ti}\in \mathbb{R}, a_i\in\mathbb{R}^{K}\\
x_t&=W_eS_t, &&t\in\{0,...,N-1\}\\
p_{t+1}&={\rm Decoder}(x_t, z_t), &&t\in\{0,...,N-1\}
\end{align}

%The ${\rm Decoder}$ is similar to the LSTM in the basic Encoder-Decoder but needs to pass the context vector $z_t$ calculated at time $t$ to the input of $i_t, f_t, o_t, g_t$.

The above Eqs is the Soft attention mechanism proposed in the paper, details are shown in Figure 3 (left), and another Hard attention is also proposed. However, most of the improved models use Soft attention easy to implement, so only the Soft attention mechanism is introduced here.

\subsection{Improvements in Encoder}
\cite{you_image_2016} proposes a semantic attention model, in addition to using CNN's intermediate activation output as the global feature of the image $v$, and also using a set of attribute detectors to extract $\{A_i\}$ the most likely to appear in the image. Each attribute $A_i$ corresponds to an entry in the vocabulary, so the model encodes the image as a collection of visual features and semantic features. Then adaptively process $\{A_i\}$ to calculate the input of the Decoder $x_t$ and get the current word output $p_t$.
\begin{align}
v&={\rm Encoder}(I)\\
h_t&={\rm Decoder}(h_{t-1},x_t)\\
p_t&=\varphi(h_t,\{A_i\})\\
x_t&=\phi(p_{t-1},\{A_i\})
\end{align}

\cite{liu_mat:_2017} returns the image captioning problem back to machine translation, which first uses the object detector to represent the image $I$ as the sequence of detection objects $seq(I)=\{ O_1, O_2,...,O_{T_A}\}$, where $\{O_1,...,O_{T_A-1}\}$ is the image objects feature representation, the last item $O_{T_A}$ is the global feature of the image; then applies the Sequence 2 Sequence framework in machine translation to $seq(I)$ to generate the image description $S=\{S_1,S_2,...,S_{T_B}\}$, Encoder and Decoder are implemented using LSTM.
\begin{align}
h_{t_E}={\rm Encoder}(O_{t_E}, h_{t_E-1}), t_E=1,2,...,T_A\\
d_{t_D}={\rm Decoder}(S_{t_D}, d_{t_D-1}), t_D=1,2,...,T_B
\end{align}
In addition, when $S_T$ is generated, the model applies the attention mechanism to generate $d_{t-1}'$ on the Encoder hidden layer sequence output $h=\{h_1, h_2, ..., h_{T_A}\}$. Then cocat $d_{t-1}'$ and $d_t$ use the Softmax activation function to generate the current $S_t$.

\cite{chen_sca-cnn:_2016} believes that CNN's kernels can be used as pattern detectors, and each channel of image feature map is activated by the corresponding convolution kernel. Therefore, the application of attention mechanism on the channel can be regarded as a process of selecting image semantic attributes. They proposed SCA-CNN, which applies the attention mechanism to both space and channel. However, unlike the previous attention mechanism, when calculating the context vector, they only weight the region features without summing, which can ensure that the feature vector and the context vector are the same sizes, so the SCA-CNN can be embedded in the stack multiple times.

\cite{fu_aligning_2017} introduced advanced semantic information to improve image description based on attention. Firstly, the object detection generates a series of candidate regions, and a two-classifier is used to classify the candidate regions (good/bad). Finally, the first 29 regions and the image global region are selected as the visual feature information. The attention mechanism generates a context vector $z_t$. In addition, they use LDA to model all descriptions in the dataset to map the images flexibly to 80-Dimensional topic vectors (corresponding to the implicit 80 scene categories) and then train a multi-layer perceptron to predict the scene context vector $s$  to better generate image descriptions.

\cite{yao_exploring_2018} believes that the semantic relationship and spatial relationship between image objects are helpful for image description generation. They first use the object detection module Faster R-CNN \cite{DBLP:conf/nips/RenHGS15} to detect objects in the image, and represent the image as $K$ image saliency area containing the object $V=\{v_i\}_{i=1}^K$; Then use a simple classification network to predict the semantic relationship between the objects and construct a semantic relationship graph $\mathcal{G}_{sem}=(V,\varepsilon_{sem})$, and construct a spatial relationship graph $\mathcal{G}_{spa}=(V, \varepsilon_{spa})$ by using the positional relationship of the object area. Then they design a [GCN]-based image Encoder to fuse the semantic and spatial relationships between the objects to obtain visual features $V^{(1)}=\{v_i^{( 1)}\}_{i=1}^K$ containing more information.

As can be seen from the above, the original intention of improving Encoder is mostly to extract more useful information from images, such as adding semantic information on the basis of visual information, replacing the original CNN response activation region with the object detection module. Therefore, these methods have improved the image description effect, but there are also some inherent defects. On one hand, object detection may affect the efficiency of image description generation, on the other hand, it is difficult to effectively interpret the reliability of the semantic information of the image acquired implicitly.

\begin{figure}[]
    \centering
    \subfigure{\includegraphics[width=3.2in]{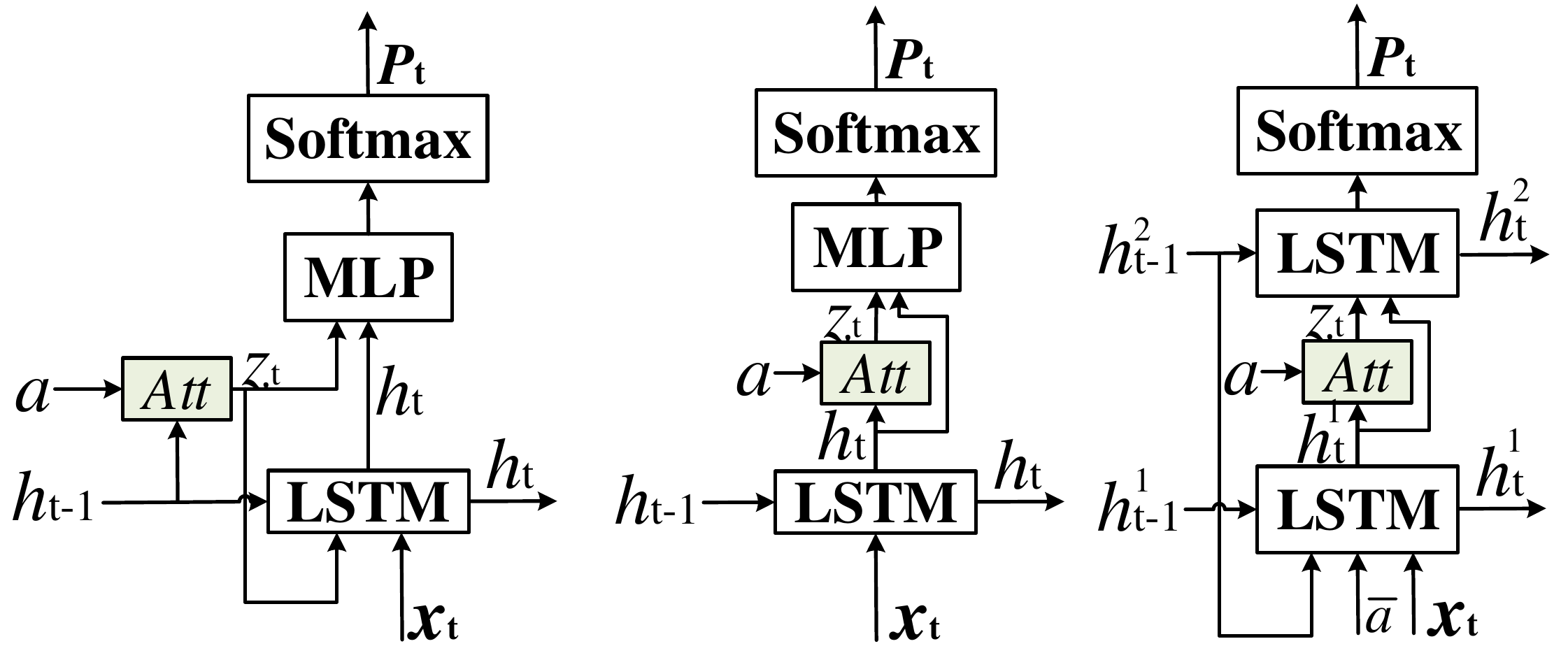}}
    \caption{Attention Language Model Details.}
\end{figure}

\subsection{Improvements in Decoder}
\cite{lu_knowing_2017} believes that in the process of generating image description, visual attention should not be added to non-visual words such as prepositions and quantifiers. So they introduced a visual sentinel in Decoder, essentially adding a gating to the LSTM for generating the sentinel vector $s_t$ for each time step. In addition, they think that visual attention should be more relevant to the current time step hidden layer state of LSTM, so the visual attention mechanism is improved compared to \cite{xu_show_2015}, see Figure 3 (center). When visual attention weights $\alpha_t$ are generated, the weight value $\beta_t$ is calculated to determine whether to visually focus on the image. Therefore the context vector for each time step is calculated as follows.
\begin{align}
\alpha_{ti}&\propto \exp\{f_{\rm Vatt}(a_i,m_{t})\}\\
\beta_t&\propto \exp\{f_{\rm Satt}(s_t,m_{t})\}\\
z_t&=\beta_t s_t+(1-\beta_t)\sum_{i=1}^L\alpha_{ti}a_i
\end{align}

\cite{anderson_bottom-up_2017} combines Bottom-Up and Top-Down attention. Firstly, based on Faster R-CNN as Bottom-Up attention model, a variable-size image feature set,$V=\{v_i\}_{i=1}^K$, is obtained. Each feature is the encoding of a salient region of the image. Decoder used to generate language descriptions uses a two-tier LSTM structure, see Figure 3 (right). The first LSTM acts as Top-Down attention layer, which applies attention mechanism on hidden layer output and visual feature $V$ to calculates context vector $z_t$. Then it is fed into the second LSTM and delivers the output of the second LSTM to Softmax classifier to generate the current time step word prediction.

\cite{zhou_watch_2017} pointed out that in previous work, image features are only initially fed into LSTM, or on the basis of which attention mechanism is introduced to compute context vectors to input LSTM. Whether text context could be used to improve image description performance has not been solved yet, that is, the relationship between generated words and visual information was not involved. To explore this problem, they proposed a Text-Conditional attention mechanism, which allows attention to focus on image features related to previously generated words. They fused the previously generated words with global image features $I$ to generate context vector $z_t$, and then input them to LSTM to generate words $S_{t+1}$.
\begin{align}
z_t&=\phi(I\odot W_CS_t)\\
z_t&=\phi(I\odot W_C\sum_{k=1}^t\frac{S_{k-1}}{t})
\end{align}
Eq.(19) is Text-Conditional attention in the form of 1-gram, and the context information is limited to the previous word; Eq.(20) is an extreme form, and the context information takes advantage of all the previously generated words.

In most work, RNN in one or two layers is used as a language model to generate descriptive words. \cite{fang_looking_2018} thinks that this structure can deal with visual words such as nouns more easily, but it may not be able to learn verbs and adjectives. Therefore, they proposed a deep attention language model based on multi-layer LSTM, which can learn more abstract word information, and design three overlapping methods to generate attention context vectors.

LSTM is often used as Decoder part in image captioning tasks, but LSTM is relatively complex and can not be performed in parallel. \cite{aneja_convolutional_2018} and \cite{wang_cnn+cnn:_2018} proposed that CNN is used as Decoder part to generate image description, which can achieve the same effect as LSTM and greatly improve the computing speed.

When using RNN (e.g. LSTM and GRU) as Decoder to generate description, Decoder's input, hidden states and output are usually expressed as 1-D vectors. \cite{dai_rethinking_2018} considers that 2-D feature mapping is more effective in interpretation and convenient for visual analysis to study the relationship between input visual information and output descriptive words; secondly, 2-D features can retain important spatial structure information. Therefore, they proposed to design Decoder on 2-D feature maps. Firstly, CNN is used to transform an image into multi-channel 2-D feature mapping. Decoder still uses GRU structure, but the state mapping transformations is replaced by convolution operations.

The above works show that the improvement of Decoder mainly focuses on the richness of information and the correctness of the attention when generating the description.

\subsection{Other Improvements}
On the basis of Encoder and Decoder, \cite{yang_review_2016} introduced a Reviewer module, which is essentially an improved LSTM unit introducing attention mechanism. It is used to perform multiple Reviews on the local features of Encoder output, and calculate a fact vector $f_t$ at each step as the input of attention module in Decoder. The author considers that the fact vector extracted by Reviewer module is more compact and abstract than the image feature maps obtained by Encoder. Therefore, the visual attention of the model is applied to the Reviewer module, while the Decoder module applies the attention mechanism to the fact vector.

Two forms of Reviwer module are introduced in this paper. One is Attention Input Reviewer, which first applies the attention mechanism to the image region features $a$ and then uses the attention output as the input of LSTM unit to generate the fact vector $f_t$,
\begin{align}
\alpha_{ti}&\propto \exp\{f_{\rm att}(a_i,f_{t-1})\}\\
\tilde{f}_{t}&=\sum_{i=1}^L\alpha_{ti}a_i\\
f_t&={\rm LSTM_{R}}(\tilde{f}_{t}, f_{t-1})
\end{align}
Another one is Attention Output Reviewer, which also applies attention mechanism to image region features, but uses zero vector as input of LSTM unit, fact vector is calculated as the sum of LSTM output and attention output,
\begin{align}
f_t&={\rm LSTM_{R}}(0, f_{t-1})+W\tilde{f}_{t}
\end{align}

Inspired by \cite{yang_review_2016}, \cite{jiang_learning_2018} designs a Guiding Network based on a simple neural network in Encoder and Decoder structure. The region feature set of the image is used as input to generate a guidance vector $v$ containing the global information of the image. The guidance vector $v$ will then be fused with the original input of the Decoder to ensure that richer image information is input when generating image descriptions.

%\begin{table}[]
%\caption{Evaluation results of partial models}
%\centering
%\begin{tabular}{lrrrrrrr}
%\toprule
%\multicolumn{1}{c}{} & \multicolumn{5}{c}{MS COCO}\\
%& B3 & B4 & MT & CIDEr & SPICE\\
%\midrule
%\cite{DBLP:conf/cvpr/VinyalsTBE15}  & 32.9 & 24.6 & - & - & -\\
%\cite{xu_show_2015} & 34.4 & 24.3 & 23.9 & - & -\\
%%\midrule
%\cite{you_image_2016} & 40.2 & 30.4 & 24.3 & - & - \\
%\cite{liu_mat:_2017} & \textbf{42.9} & 32.3 & 25.8 & 105.8 & 18.9\\
%\cite{chen_sca-cnn:_2016} & 41.1 & 31.1 & 25.0 & - & -\\
%\cite{fu_aligning_2017} & 41.8 & 31.3 & 24.8 & 95.5 & -\\
%\cite{yao_exploring_2018} & - & \underline{\textbf{37.1}} & \underline{\textbf{28.1}} & \underline{\textbf{117.1}} & \underline{\textbf{21.1}}\\
%%\midrule
%\cite{lu_knowing_2017} & 43.9 & 33.2 & 26.6 & 108.5 & -\\
%\cite{anderson_bottom-up_2017} & - & \textbf{36.2} & \textbf{27.0} & \textbf{113.5} & \textbf{20.3}\\
%\cite{zhou_watch_2017} & 40.5 & 30.1 & 24.7 & 97.0 & -\\
%\cite{fang_looking_2018} & \underline{\textbf{44.2}} & 34.0 & 26.4 & 105.6 & -\\
%\cite{aneja_convolutional_2018} & 39.4 & 28.7 & 24.4 & 91.2 & 17.5\\
%\cite{wang_cnn+cnn:_2018} & 36.9 & 26.7 & 23.4 & 84.4 & -\\
%\cite{dai_rethinking_2018} - & 31.9 & - & 99.4 & 18.7\\
%\bottomrule
%\end{tabular}
%\end{table}

\begin{table*}[]
\centering
\begin{tabular}{lrrrrrrrrrrrrr}
\toprule
\multicolumn{1}{c}{} & \multicolumn{6}{c}{Flickr30K} & \multicolumn{7}{c}{MS COCO}\\
Methods & B-1 & B-2 & B-3 & B-4 & MT & CD & B-1 & B-2 & B-3 & B-4 & MT & CD & SP\\
\midrule
\cite{DBLP:conf/cvpr/VinyalsTBE15}  & 66.3 & 42.3 & 27.7 & 18.3 & - & - & 66.6 & 46.1 & 32.9 & 24.6 & - & - & -\\
\cite{xu_show_2015} & 66.7 & 43.4 & 28.8 & 19.1 & 18.49 & - & 70.7 & 49.2 & 34.4 & 24.3 & 23.9 & - & -\\
%\midrule
\cite{you_image_2016} & 64.7 & 46.0 & 32.4 & 23.0 & 18.9 & - & 70.9 & 53.7 & 40.2 & 30.4 & 24.3 & - & - \\
\cite{liu_mat:_2017} & - & - & - & - & - & - & 73.1 & {56.7} & {42.9} & 32.3 & 25.8 & 105.8 & 18.9\\
\cite{chen_sca-cnn:_2016} & {66.2} & {46.8} & {32.5} & 22.3 & {19.5} & - & 71.9 & 54.8 & 41.1 & 31.1 & 25.0 & - & -\\
\cite{fu_aligning_2017} & 64.9 & 46.2 & 32.4 & {22.4} & 19.4 & {47.2} & 72.4 & 55.5 & 41.8 & 31.3 & 24.8 & 95.5 & -\\
\cite{yao_exploring_2018} & - & - & - & - & - & - & {{77.4}} & - & - & {37.1} & {28.1} & {117.1} & {21.1}\\
%\midrule
\cite{anderson_bottom-up_2017} & - & - & - & - & - & - & {77.2} & - & - & {36.2} & {27.0} & {113.5} & {20.3}\\
\cite{zhou_watch_2017} & - & - & - & - & - & - & 71.6 & 54.5 & 40.5 & 30.1 & 24.7 & 97.0 & -\\
\cite{fang_looking_2018} & - & - & 32.8 & 23.4 & 18.7 & 43.7 & - & - & {44.2} & 34.0 & 26.4 & 105.6 & -\\
\cite{aneja_convolutional_2018} & - & - & - & - & - & - & 71.1 & 53.8 & 39.4 & 28.7 & 24.4 & 91.2 & 17.5\\
\cite{wang_cnn+cnn:_2018} & 60.7 & 42.5 & 29.2 & 19.9 & 19.1 & 39.5 & 68.5 & 51.1 & 36.9 & 26.7 & 23.4 & 84.4 & -\\
\cite{dai_rethinking_2018} & - & - & - & 22.0 & - & 42.7 & - & - & - & 31.9 & - & 99.4 & 18.7\\
\bottomrule
\end{tabular}
\caption{Evaluation results of some models. B-n, MT, CD, SP stand for BLEU-n, METEOR, CIDEr and SPICE respectively.}
\label{tab:booktabs}
\end{table*}

\section{Datasets}
Image captioning based on deep learning methods requires a lot of label data. Fortunately, many researchers and research organization have collected and tagged data sets. Here we mainly introduce four common data sets: Flickr 8K \cite{hodosh_framing_2013}, Flickr 30K \cite{young_image_2014}, MS COCO \cite{lin_microsoft_2014} and Visual Genome\cite{krishna_visual_2017}.

\textbf{Flickr8K} \cite{hodosh_framing_2013} contains a total of 8092 images, which were collected from Flickr.com and captions were obtained through crowdsourcing services provided by Amazon Mechanical Turk. Each image contains five different captions for reference with an average length of 11.8 words, and these descriptions are required to accurately describe the objects, scenes, and activities displayed in the image. In practical applications, 8000 images are usually selected, of which 6000 for train, 1000 for verification, and 1000 for test.

\textbf{Flickr30K} \cite{young_image_2014} is an extension to Flickr8K. It contains 31,783 images (including 8092 images in Flickr8K) and 158,915 descriptions. An annotation guide similar to Flickr8K is used to obtain image descriptions, control description quality, and correct description errors. Usually, 1000 images are selected as validation data, 1000 images as test data, and the remaining images are used as train data.

\textbf{MicroSoft COCO} \cite{lin_microsoft_2014} is a large-scale dataset that can be used for object detection, instance segmentation, and image captioning. It is also the most popular dataset in image captioning. The dataset contains 91 object categories, a total of 328K images, 2.5 million tag instances, and each image contains 5 descriptions. The dataset is divided into two parts. The part released in 2014 includes 82,783 train data, 40,504 validation data and 40,775 test data. However, the description of the test set is not publicly available, so the train set data and the validation set data are often re-divided into training/validation/test set in practical applications.
%The 2015 release includes 165,482 training set images and 81,208 validation set images. 81,434 test set images. 

%\subsection{Visual Genome}
\textbf{Visual Genome} \cite{krishna_visual_2017} contains more than 108K images. Each image contains an average of 35 objects with dense description annotations, 26 attributes and 21 interactions between objects. Therefore, Visual Genome dataset can be used to pre-train image captioning tasks that introduce spatial and semantic relationships between objects.
%is used for pre-training in some methods that use object detection module as Encoder. Visual Genome

\section{Evaluation}
\textbf{BLEU}\cite{papineni_bleu:_2002} is the most commonly used evaluation metric in image captioning tasks. It was originally used to measure the quality of machine translation. The core idea of BLEU is that "the closer the test sentences are to the reference sentences, the better". In other words, BLEU is evaluated by comparing the similarity of the test sentences and the reference sentences at the n-gram level. Therefore, this method does not consider the grammatical correctness, synonyms, similar expressions, and is more credible only in the case of shorter sentences.\\
\textbf{METEOR} \cite{banerjee_meteor:_2005} is also a commonly used evaluation metric for machine translation. Firstly, test sentences are aligned with reference sentences, such as word precise matching, stemmer-based matching, synonym matching and alignment based on WordNet, etc. Then, similarity scores between the test and the reference sentences are calculated based on alignment results. The calculation of similarity scores involves such indicators as matching word accuracy and recall rate. This method solves some shortcomings of BLEU and can express better relevance at the sentence level.\\
\textbf{CIDEr} \cite{vedantam_cider:_2015} is an evaluation metric aiming at image captioning. The authors think that the past evaluation metrics have a strong correlation with human, but they can not evaluate the similarity between them and human. So they proposed Consensus-based evaluation metric. Each sentence is regarded as a "document" and expressed as a TF-IDF vector. The weight of TF-IDF is calculated for each n-gram, and then the cosine similarity between the test sentences and the reference sentences is calculated for evaluation.\\
\textbf{SPICE} \cite{anderson_spice:_2016} is also an evaluation metric designed for image captioning. The metric codes the objects, attributes and relationships in image description into a semantic graph. This method captures the human's judgment of model generation description better than the existing n-gram based evaluation metrics and can reflect the advantages and disadvantages of the language model more accurately.

\section{Discussion \& Future Research Directions}
The evaluation results of some deep learning methods are shown in Table 1, which shows that deep learning methods have achieved great success in image captioning tasks. In the previous part, we mainly discussed the improved model based on Encoder-Decoder structure. The emphasis of different improvements is different, but most of them aim to enrich the visual feature information of images, which is also a common original intention of them. For example, the improvement of Encoder includes extracting more accurate salient region features from images by object detection, enriching visual information of images by extracting semantic relations between salient objects from images, and implicitly extracting a scene vector from images to guide the generation of descriptions, all of which are in order to obtain richer and more abstract information from images or obtain additional information. Further improvements of Decoder include increasing the use of previously generated descriptive words, adding control gates to language models to ensure proper application of attention mechanisms, and implicitly increasing the number of layers of LSTM to obtain more abstract information.

Nevertheless, image captioning is far from the human level, so there is still much space for improvement. On one hand, we can continue to study how to extract richer visual information from images or combine the extracted feature maps into more abstract information to enhance the context features of Decoder. Such as introducing semantic segmentation into Encoder part and using the latest language models as Decoder; on the other hand, I think we can deepen the development of datasets. Existing image captioning datasets only correspond images and descriptions, regions of interest of descriptions and how to generate descriptions are not reflected. If the development of datasets can be strengthened, more monitoring information can be introduced into the training of models, which may improve the performance of image captioning.

\section{Conclusion}
In this paper, image captioning based on deep learning methods is summarized. Firstly, traditional template-based and retrieval-based methods are briefly introduced. Secondly, the deep learning methods and their improvements based on Encoder-Decoder structure in recent years are mainly introduced. According to the emphasis on improvements, these improvements are divided into three parts: Encoder Improvements, Decoder Improvements, and Other Improvements. Then, we introduce the commonly used datasets and evaluation metrics in image captioning. Although image captioning based on deep learning has been improved, they also have much space for improvements. So finally, we summarize the results of some deep learning methods and forecast future research directions. 

%% The file named.bst is a bibliography style file for BibTeX 0.99c
\bibliographystyle{named}
\bibliography{ijcai19}
\end{CJK}
\end{document}